# A Novel adaptive optimization of Dual-Tree Complex Wavelet Transform for Medical Image Fusion

[A] G.Karpaga Kannan, [B] T.Deepika.

[A] Assistant Professor, [B] PG Student.

[A, B] RVS College of Engineering and Technology, Dindigul, Tamilnadu

[A] karpagakannan@gmail.com, [B] deepikarvs@gmail.com.

**Abstract-** In recent years, many research achievements are made in the medical image fusion field. Fusion is basically extraction of best of inputs and conveying it to the output. Medical Image fusion means that several of various modality image information are comprehended together to form one image to express its information. The aim of image fusion is to integrate complementary and redundant information. In this paper, a multimodal image fusion algorithm based on dual tree complex wavelet transform (DT-CWT) and adaptive particle swarm optimization (APSO) is proposed. Fusion is achieved through the formation of a fused pyramid using the DTCWT coefficients from the decomposed pyramids of the source images. The coefficients are fused by weighted average method based on pixels, and the weights are estimated by the APSO to gain optimal fused images. The fused image is obtained through conventional inverse dual tree complex wavelet transform reconstruction process. Experiment results show that the proposed method based on adaptive particle swarm optimization algorithm is remarkably better than the method based on particle swarm optimization. The resulting fused images are compared visually and through benchmarks such as Entropy (E), Peak Signal to Noise Ratio, (PSNR), Root Mean Square Error (RMSE), Standard deviation (SD) and Structure Similarity Index Metric (SSIM) computations.

## I. INTRODUCTION

The fast development of digital image processing leads to the growth of feature extraction of images which leads to the development of Image fusion. Image fusion is that two or more images of the same object which collected from multi-source image sensors synthesis in one image that is more suitable for human visual or machine perception [6]. Recently, in order to better support more accurate clinical information for physicians to deal with medical diagnosis and evaluation, multimodality medical images are needed, such as X-ray, computed tomography (CT), magnetic resonance imaging (MRI), and positron emission tomography (PET) images, etc [7]. For medical image fusion, the fusion of images can often lead to additional clinical information not apparent in the separate images. Another advantage is that it can reduce the storage cost by storing just the single fused image instead of multi-source images [2]. However, due to the real-world objects usually contain structures at many different scales or resolutions, the Multiresolution techniques have then attracted more and more interest in image fusion [1].

In the last decade, many software and techniques have been developed to resolve the image fusion problem. In all those methods, the fusion schemes based on Multiresolution transform have attracted a considerable amount of research attention. Some popular transforms include discrete wavelet (DWT) [14], stationary wavelet (SWT), dual-tree complex wavelet (DTCWT) [15], curvelet (CVT), Contourlet (CT) and Nonsubsampled Contourlet transform (NSCT). In addition, Zheng et al. in [16] proposed an image fusion method based on the support value transform, which used the support value to represent the salient features of image.

J. Kennedy and R. C. Eberhart brought forward particle swarm optimization (PSO) inspired by the choreography of a bird flock in 1995 [5]. Unlike conventional evolutionary algorithms, PSO possesses the following characteristics: 1) Each individual (or particle) is given a random speed and flows in the decision space; 2) each individual has its own memory; 3) the evolutionary of each individual is composed of the cooperation and competition among these particles. It has been of great concern and become a new research field. In order to improve the performances of the PSO algorithm, we present a proposal, called "adaptive particle swarm optimization (APSO)" is presented in this work. APSO algorithm is implemented to optimize the objective function, because of its high speed and its excellent performance in global search. In contrast to other evolutionary algorithms, APSO has a higher convergence speed and better exploratory capabilities. The approach to image fusion based on APSO is more successful.

In this paper, a new multi objective optimization method of multi objective image fusion based on APSO (Adaptive Particle Swarm Optimization) is proposed based on pixel level, which can simplify the model of multi objective image fusion and overcome the limitations of traditional methods. First the proper evaluation indices of multi objective image fusion are given, then the uniform model of multi objective image fusion in DT-CWT domain is constructed, in which the model parameters are selected as, the decision variables, and finally APSO is designed to optimize the decision variables. APSO not only uses a mutation operator to avoid earlier convergence, but also uses a crowding operator to improve the distribution of no dominated solutions and uses a new adaptive inertia weight to raise the optimization capacities.

A DTCWT based fusion scheme with APSO is proposed to produce the optimal fused result adaptively. The source images are firstly decomposed into low-frequency and high-frequency coefficients by DTCWT. For the fusion of coefficients, to highlight different parts adaptively, we fuse the coefficients with a pixel based weighted average fusion rule. The weights are optimized with APSO. The fused image is reconstructed with fused low-frequency and high-frequency coefficients. The fusion algorithm proposed, which can avoid the local optimum in APSO. Experimental results on two real images show that the method based on PSO is better than that based on APSO and can improve performance further.

The rest of this paper is organized as follows. In Section 2, overview of the DTCWT based image fusion scheme is given. Section 3 and 4 provides the procedure of the proposed method followed by some benchmarks evaluation and experiments in Section 5 & 6 and conclusion in Section 7.

## II. A BRIEF OVERVIEW OF DTCWT

The dual-tree complex wavelet transform is recent enhancements to the discrete wavelet transform [4]. The transform was first proposed by Kingsbury [9] in order to mitigate two main disadvantages, namely, the lack of shift invariance and poor directional selectivity, of the discrete wavelet transform (DWT) [3]. There are two versions of the 2D DTWT transform namely Dual Tree Discrete Wavelet Transform (DTDWT) which is 2-times expansive, and Dual Tree Complex Wavelet Transform (DTCWT) which is 4-times expansive [8]. Dual Tree Complex Wavelet Transform, a form of discrete wavelet transform which generates complex coefficients by using a dual tree of wavelet filters to obtain their real and imaginary parts. The properties of the DT-CWT can be summarized as [11]

a) Approximate shift invariance;
b) Good directional selectivity in 2 dimensions;
c) Phase information;
d) Perfect reconstruction using short linear-phase filters;
e) Limited redundancy, independent of the number of scales, 2: 1 for 1D ($2m:$ 1 for $m$D);
f) Efficient order-N computation—only twice the simple DWT for 1D (2m times for mD).

Decompose the image with DT-CWT; it produces two sub bands of the low frequency and six sub bands of different directions in the high frequency at each level of decomposition, where each high-pass sub band corresponds to one unique direction θ [12].

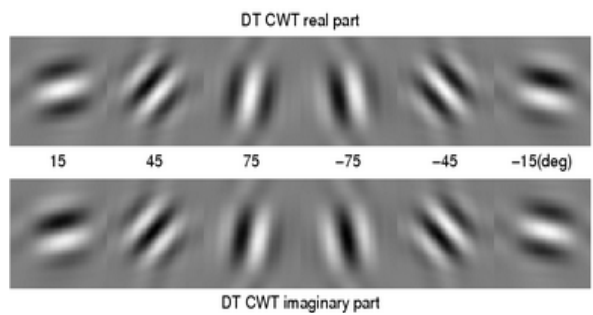

Fig 1: The real and imaginary impulse responses of the DTCWT.

It can help to improve the precision of the decomposition and the reconstruction of the images. It covers more distinct orientations than the separable DWT wavelets and has the ability to differentiate positive and negative frequencies. Fig. 1 shows the impulse responses of the dual-tree complex wavelets.

Fig 2 shows the DT-CWT structure includes two parallel discrete wavelets as two trees, which create the real part and the imaginary part of the wavelet coefficients. The two trees use different filters, in the process of the decomposition of the first layer, choose odd lengths filters. For the decomposition of the other layers, the lengths of one tree's filters are odd and the other one's are even. For different levels of each tree, the two trees will present the good symmetry when use the odd and even filters alternately.

One way to double all the sampling rates in a conventional wavelet tree is to eliminate the down-sampling by 2 after the level 1 filter. This is equivalent to having two parallel fully-decimated trees, provided that the delays of filters are one sample offset from the delays, which ensures that the level 1 downsamplers in tree *b* pick the opposite samples to those in tree *a*. Then find that, to get uniform intervals between samples from the two trees below level 1, the filters in one tree must provide delays that are half a sample different from those in the opposite tree. For linear phase filters, this requires odd-length filters in one tree and even-length filters in the other.

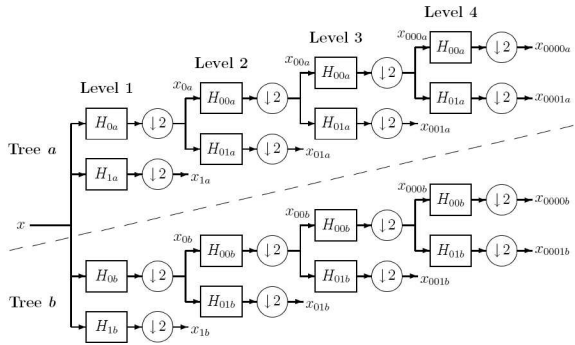

Fig 2: The DT-CWT structure

The DT-CWT also yields perfect reconstruction by using two parallel decimated trees with real-valued coefficients generated at each tree. The inverse of the dual-tree CWT is as simple as the forward transform. To invert the transform, the real part and the imaginary part are each inverted—the inverse of each of the two real DWTs are used—to obtain two real signals[10]. These two real signals are then averaged to obtain the final output.

### III. DT-CWT BASED IMAGE FUSION PROCEDURE

The approach to image fusion based on multi-model optimization in DT-CWT domain is as follows. The fusion scheme with two input images I1 and I2 which are assumed to have been registered. Firstly, each of the registered input images is transformed to low frequency coefficients A and series of high-frequency coefficients D by applying DTCWT, defined as:

$$(A, D) = DTCWT (I). \quad (1)$$

Then, in general, the low-frequency coefficients $A_1$, $A_2$ and high-frequency coefficients $D_1$, $D_2$ are handled separately to give fused version, defined as:

$$A_F = \Phi_A (A_1, A_2)$$
$$D_F = \Phi_D (D_1, D_2) \quad (2)$$

where $\Phi_A$, $\Phi_D$ are fusion rules for low-frequency and high-frequency coefficients, and $A_F$, $D_F$ indicate the fused low-frequency and high-frequency coefficients. Finally, the fused image F is obtained by performing the inverse transform (IDTCWT) on $A_F$ and $D_F$, defined as:

$$F = IDTCWT (A_F, D_F) \quad (3).$$

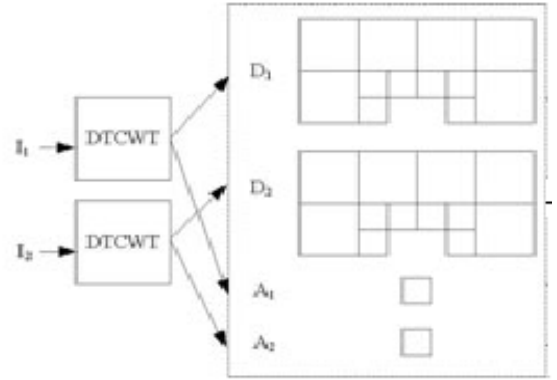

Fig 3: The fusion scheme based on DTCWT

### IV. THE PROPOSED FUSION TECHNIQUE

After decomposing input images into low-frequency and high-frequency coefficients, in order to highlight region features of the source images, the corresponding coefficients are fused and the values of weights are adaptively estimated by APSO with the fitness function, is proposed. In this paper, all the APSO algorithms studied are based on the gbest model, aiming to reduce the probability of being trapped into local optima by introducing mutation & crowding operator and keep the advantage of fast con-vengeance of the gbest model.

The flow of APSO is as follows. Initialize the population and algorithm parameters.

1) Initialize the position of each particle: pop[i], where i=1,…, NP, NP is the particle number.
2) Initialize the speed of each particle: vel [i] =0
3) Initialize the record of each particle: pbest[i] = pop[i]
4) Evaluate each of the particles in the POP: fun [i, j], where j=1,…, NF, and NF is the objective number.
5) Store the positions that represent no dominated particles in the repository of the REP according to the Pareto optimality.

Before the maximum number of cycles is reached, do
1) Update the speed of each particle using (4).

$$Vel[i] = W \cdot Vel[i] + C_1 \cdot rand_1 \cdot (Pbest[i] - pop[i]) + C_2 \cdot rand_2 \cdot (Gbest[i] - pop[i]) \quad (4)$$

where W is the inertia weight [18]; c1 and c2 are the learning factors [19], rand1 and rand2 are random values in the range [0, 1]; pbest[i] is the best position that the particle i has had; h is the index of the maximum crowding distance in the repository that implies the particle locates in the sparse region, as aims to maintain the population diversity; pop[i] is the current position of the particle i.
2) Update the new positions of the particles adding the speed produced from the previous step

$$Pop[i] = pop[i] + vel[i] \quad (5)$$

3) Maintain the particles within the search space in case they go beyond their boundaries. When a decision variable goes beyond its boundaries, the decision variable takes the value of its corresponding boundary, and its velocity is multiplied by (-1).
4) Adaptively mutate each of the particles in the POP at a probability of Pm.
5) Evaluate each of the particles in the POP.
6) Update the contents in the REP, and insert all the current no dominated positions into the repository.
7) Update the records, when the current position of the particle is better than the position contained in its memory, the particle's position is updated.

$$Pbest[i] = pop[i] \quad (6)$$

8) Increase the loop counter of g.

For approximations in DT-CWT domain, use weighted factors to calculate the approximation of the fused image of F. Using multi-modal optimization methods, find the optimal decision variables of image fusion in DT-CWT domain, and realize the optimal image fusion. The fused image F is reconstructed by the fused low-frequency coefficients $A_F$ and the fused high frequency coefficients $D_F$. The new sets of coefficients are used to find the inverse transform to get the fused image F.

## V. QUALITY EVALUATION

The quality of the fused image is evaluated by five benchmarks: Entropy (E), Root Mean Square Error (RMSE), Peak-to-Peak Signal-to-Noise Ratio (PSNR), Structure Similarity Index Metric (SSIM) and Standard deviation (SD). Consider R as the source image and F the fused image, both of size M×N. F (i, j) is the grey value of pixel at the position (i, j).

1) Entropy (E)
Entropy is an index to evaluate the information quantity contained in an image. If the value of entropy becomes higher after fusing, it indicates that the information increases and the fusion performance are improved.

$$E = -\sum_{i=0}^{L-1} p_i \log_2 p_i$$

Entropy is defined as where L is the total of grey levels, p= {p0, p1…pL-1} is the probability distribution of each level [13].

2) Peak-to-Peak Signal-to-Noise Ratio (PSNR)
PSNR is the ratio between the maximum possible power of a signal and the power of corrupting noise that affects the fidelity of its representation. The PSNR measure is given by [13]

$$PSNR = 10 \log 10 \, (255)^2 / (RMSE)^2 \, (db)$$

3) Root Mean Square error
The root Mean Square error is defined as follows:

$$RMSE = \sqrt{\frac{\sum_{i=1}^{M} \sum_{j=1}^{N} [R(i,j) - F(i,j)]^2}{M \times N}}$$

The RMSE is used to measure the difference between the source image and the fused image; the smaller the value of RMSE and the smaller the difference, the better the fusion performance.

4) Structure Similarity Index Metric (SSIM)
SSIM measures the structural similarity between the fused and the reference images.

$$SSIM = \frac{(2\mu_f \mu_r + C_1)(2\sigma_f \sigma_r + C_2)}{(\mu^2_f + \mu^2_r + C_1)(\sigma^2_f + \sigma^2_r + C_2)}$$

5) Standard deviation (SD)

Standard deviation is shown as follows

$$SD = \sqrt{\frac{1}{MXN}\sum_{m=1}^{M}\sum_{n=1}^{N}(F(m,n)-MEAN)^2}$$

Where MEAN is the average denoted by

$$MEAN = \frac{1}{MXN}\sum_{m=1}^{M}\sum_{n=1}^{N}|F(m,n)|$$

## VI. EXPERIMENTAL RESULTS

The performance of the proposed image fusion approach was tested and compared with that of PSO scheme. Proposed algorithm is applied to multimodality medical images (CT and MR images). The parameters of APSO are as follow: the particle number of NP is 100; the objective number of NF is 6; the inertia weight of W is 0.5; the learning factor of $C_1$ is 1, and C2 is 1; the maximum cycle number of $G_{max}$ is 100; the allowed maximum capacity of MEM is 100; the mutation probability of Pm is 0.05. Multimodality CT and MR medical images are fused and shown in figure 4. A fused image using the proposed method is found to have less ringing artifacts than the DWT based modulus maxima approach. The inverse wavelet transform is the fused image with clear focus on the whole image. Their performance is measured in terms of Entropy, Peak-to-Peak Signal-to-Noise Ratio, Root mean square errors, structure similarity index metric & Standard deviation and tabulated in table1 & 2. The visual experiments and the quantitative analysis demonstrate that the proposed medical image fusion method can preserve the important structure information such as edges of organs, outlines of images compared to other image fusion methods.

## VII. CONCLUSION

This paper describes a multimodal image fusion algorithm based on DTCWT and PSO to obtain optimal fused images. For fusion of the multimodal images of different intensity range, it is vital to adjust the different intensity range between the input images and highlight different parts of the image. Proposed algorithm is applied to multimodality medical images (CT and MR images) along with DT-CWT. Fused image obtained using Proposed APSO based DTCWT is compared with the fused image obtained using PSO based DTCWT approach . The approach using APSO to optimize the model parameters of image fusion is feasible and relatively effective, and can get the Pareto optimal fusion result. The DTCWT and APSO method is used to retain edge information without significant ringing artifacts. It has the further advantages that the phase information is available for analysis. The proposed fusion technique provides computationally efficient and better qualitative and quantitative results. Experimental results demonstrate that the proposed fusion method outperforms some well-known methods both visually and quantitatively.

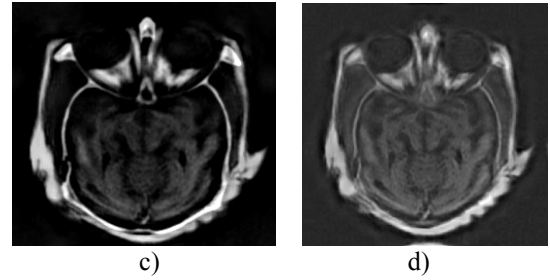

c)          d)

Fig 4:a. CT image b. MRI image c. Fused Image using DT-CWT and PSO d. Fused Image using DT-CWT and APSO.

**Table 1:** Evaluation results of the four different Benchmarks

| Benchmarks | EN | PSNR | RMSE | SD |
|---|---|---|---|---|
| CT Image | 0.7355 | 23.7999 | 27.0762 | 38.975 |
| MRI Image | 6.6325 | 12.8311 | 3.3882 | 12.364 |
| Fused using DT-CWT & PSO | 5.3091 | 15.5437 | 1.9892 | 32.455 |
| Fused using DT-CWT & APSO | 5.6551 | 16.3574 | 1.5043 | 34.689 |

**Table 2:** Evaluation results for SSIM

| Benchmark | DT-CWT & PSO | DT-CWT & APSO |
|---|---|---|
| CT & Fused Image using SSIM | 0.2675 | 0.3665 |
| MRI & Fused Image using SSIM | 0.3946 | 0.4748 |

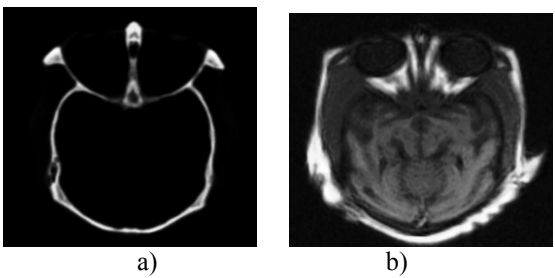

a)          b)